# Detecting Volunteer Cotton Plants in a Corn Field with Deep Learning on UAV Remote-Sensing Imagery

Pappu Kumar Yadav[a], J. Alex Thomasson[b*], Robert Hardin[a], Stephen W. Searcy[a], Ulisses Braga-Neto[c], Sorin C. Popescu[d], Daniel E Martin[e], Roberto Rodriguez[f], Karem Meza[g], Juan Enciso[a], Jorge Solorzano Diaz[h], Tianyi Wang[i]

*Corresponding Author: J. Alex Thomasson (athomasson@abe.msstate.edu, +1-662-325-3282)

150 J. Charles Lee Agricultural & Biological Engineering Building

130 Creelman Street, Mississippi State, MS 39762

[a] *Department of Biological & Agricultural Engineering, Texas A&M University, College Station, TX*

[b] *Department of Agricultural & Biological Engineering, Mississippi State University, Mississippi State, MS*

[c] *Department of Electrical & Computer Engineering, Texas A&M University, College Station, TX*

[d] *Department of Ecology & Conservation Biology, Texas A&M University, College Station, TX*

[e] *Aerial Application Technology Research, U.S.D.A. Agriculture Research Service, College Station, TX*

[f] *U.S.D.A.- APHIS PPQ S&T, Mission Lab, Edinburg, TX*

[g] *Department of Civil & Environmental Engineering, Utah State University, Logan, UT*

[h] *Texas A&M AgriLife Research & Extension Center, Weslaco, TX*

[i] *Texas A&M AgriLife Research & Extension Center, Dallas, TX*

## ABSTRACT

The cotton boll weevil, *Anthonomus grandis* Boheman is a serious pest to the U.S. cotton industry that has cost more than 16 billion USD in damages since it entered the United States from Mexico in the late 1800s. This pest has been nearly eradicated; however, southern part of Texas still faces this issue and is always prone to the pest reinfestation each year due to its sub-tropical climate where cotton plants can grow year-round. Feral or volunteer cotton (VC) plants growing in the fields of inter-seasonal crops, like corn during the winter, can serve as hosts to these pests once they reach pin-head square stage (5-6 leaf stage) and therefore need to be detected, located, and destroyed or sprayed . In this paper, we present a study to detect VC plants in a corn field using YOLOv3 convolution neural network (CNN) on three band aerial images collected by unmanned aircraft system (UAS). The two-fold objectives of this paper were : (i) to determine whether YOLOv3 can be used for VC detection in a corn field using RGB (red, green, and blue) aerial images collected by UAS and (ii) to investigate the behavior of YOLOv3 on images at three different scales (320 x 320, S1; 416 x 416, S2; and 512 x 512, S3 pixels) based on average precision (AP), mean average precision (mAP) and F1-score at 95% confidence level.  No significant differences existed for mAP among the three scales, while a significant difference was found for AP between S1 and S3 ($p = 0.04$) and S2 and S3 ($p = 0.02$). A significant difference was also found for F1-score between S2 and S3 ($p = 0.02$). The overall goal of this study was to minimize the boll weevil pest infestation by maximizing the true positive detection of VC plants in a corn field which is represented by the mAP values. The lack of significant differences of these at all the three scales indicated that the trained YOLOv3 model can be used on a computer vision-based remotely piloted aerial application system (RPAAS) for VC detection and spray application in near real-time. This has the potential to speed up the mitigation efforts for boll weevil pest control irrespective of the three input image sizes.

**Keywords**

*Boll Weevil, Volunteer Cotton Plant Detection,  Convolution Neural Network (CNN), YOLOv3, Unmanned Aircraft System (UAS), Remotely Piloted Aerial Application System (RPAAS)*

**1.0 Introduction**

The boll weevil (*Anthonomus grandis* Boheman) is an insect pest that by 2009 had cost cotton producers more than 16 billion USD (91 billion USD with consumer-price-index-adjusted value) since it entered the U.S. near Brownsville, Texas, around 1892 (Foster, 2009). To tackle the severity of boll weevil infestations, the National Cotton Council (NCC) along with USDA's Animal and Plant Health Inspection Service (APHIS) established the National Boll Weevil Eradication Program (NBWEP) in 1978, with Texas joining in the mid-1990s (McCorkle et al., 2010). This program has successfully eradicated boll weevil from most of the U.S., but in the southernmost parts of Texas, the cotton industry still accrued annual losses in the range of 150-300 million USD as of the late 1990s (Smith, 1998). The eradication program is still actively functioning in Texas to take necessary steps to prevent recurrence of these pests at larger scales and to minimize associated losses. The Texas Boll Weevil Eradication Program (TBWEP) has divided the state into 17 eradication zones, three of which are still active: Lower Rio Grande Valley (LRGV), East Texas Maintenance Area (ETMA) and West Texas Maintenance Area (WTMA). Texas Boll Weevil Eradication Foundation (TBWEF) has divided the entire state of Texas based on quarantine status. Based on this division, WTMA and parts of ETMA belong to the eradicated zone, while some regions of ETMA belong to the functionally eradicated zone, and the entire LRGV belongs to the quarantined zone (Texas Boll Weevil Eradication Foundation, Inc., 2020). The LRGV region is more prone to boll weevil infestation each year for multiple reasons. This region has a subtropical climate with hot and humid summers and mild to cool winters without freezing conditions. Cotton plants can thus grow and produce fruit year-round, on which boll weevils can feed and reproduce if not destroyed. Tropical storms coming from the Gulf of Mexico commonly carry boll-weevils from Tamaulipas State in Mexico to the LRGV (National Cotton

Council of America, 2012). This situation has made successful execution of the eradication program difficult in the LRGV region.

*1.1 Economic Impact of NBWEP*

Researchers have used various economic models to study the socio-economic impact of NBWEP in the predominant cotton growing region (Cotton Belt) of the southern U.S. Carlson and Hamming (1989) studied the economic impact of NBWEP in North and South Carolina between 1978 and 1987 and found that the program saved growers about $78 per acre per year. In another study in the Texas Rolling Plains, Robinson and Vergara (1999) found that the cost to the farmer of supporting NBWEP was less than the cost of making control applications on their own. Carlson et al. (1989) studied the cost-benefit relationship of NBWEP in the southeastern U.S. and found positive effects on cotton yield, cost savings and land values. Similarly, in the findings of Tribble et al. (1999), NBWEP in Georgia resulted in average net producer benefit of $88.73 per acre per year. The overall economic benefit of implementing NBWEP in Texas between 1996 and 2007 was close to 1.4 billion USD, while farmers' net benefits increased from $13.15 to $46.15 per acre (McCorkle et al., 2010). Szmedra et al. (1991) showed NBWEP to be a cost effective, risk efficient and net profitable program from a grower's perspective. All these studies showed net positive effects of NBWEP that need to be maintained to prevent reinfestation of boll weevil and give continuity to the benefits that result from it.

The success of NBWEP is based on three operational components: mapping, detection and control (Grefenstette and El-Lissy, 2003). In TBWEP, mapping includes all fields planted with cotton in the last two years in quarantined and functionally eradicated areas of Texas and assigning a unique identification number to each of them (Texas Boll Weevil Eradictaion Foundation Inc., 2020). Detection involves detecting volunteer cotton (VC) plants and the presence of boll weevils.

Control involves either eradication or treatment of VC plants to prevent their use as hosts by boll weevils. The control aspect first involves destroying the plants with herbicides before they reach the pin-head square phase (5 to 6 leaf stage). After VC plants reach this stage, they are difficult to kill with herbicides and therefore remain hidden in fields growing along with other crop plants (Morgan et al., 2019). Such VC plants become potential hosts to boll weevil.

*1.2 Detection and Control*

Cotton is often planted in rotation with corn or sorghum, and VC plants sometimes regrow in the subsequent crop from seeds inadvertently left in the field at harvest (Wang et al., 2022). Identifying these VC plants is difficult because they tend to blend in visually with the surrounding crop plants and may remain hidden under their canopy. VC plants are managed during five important times in a year, the growing season being the most crucial. During this season, herbicides are applied first to destroy VC plants before they reach pin-head square phase. However, most cotton varieties (more than 95%) in Texas are herbicide tolerant or resistant. Therefore, many VC plants remain alive and continue to grow beyond the pin-head square phase, making them potential hosts to boll weevil. Pesticides are thus sprayed as per the TBWEP protocol to mitigate the pest when VC plants reach pin-head square phase and beyond (Morgan et al., 2011). The study reported herein focuses on detection of VC plants before the pin-head square phase, in an early growth corn field. Detecting VC plants before the pin-head square phase has two potential benefits: aiding in precision spraying of herbicides only on the VC plants and enabling pesticides to be sprayed at those earlier-identified locations as per the TBWEP protocol after the pin-head square phase.

Mapped cotton fields are inspected for the presence of boll weevil for two consecutive years, even if they are in rotation with other annual crops. Pheromone traps are installed at outer edges of fields with cotton, fields with rotation crops, and fields adjacent to these. The traps are placed with a spacing of roughly 160 m around the perimeter of fields with cotton and with a spacing of roughly 80 m alongside overwintering habitats like shrubs, weeds, grass, etc. For the fields with rotation crops, 1 trap is placed on each side of the field, and the field is inspected first for the presence of VC plants (Fig. 1A) along the perimeters and then for the presence of boll weevils in the traps. Perimeters of the fields with rotation crops are usually inspected multiple times during the first 5 to 6 weeks in the eradicated areas of Texas, but every week in the LRGV area because of its higher susceptibility to pest infestation. When VC plants are found at the perimeters, trap spacing is increased from 1 trap per side to 1 trap every 160 m.

Cultural, mechanical, and chemical control methods are deployed in controlling boll weevils. Chemical control is most widely used, for which malathion ULV (FYFANON® ULV AG) is sprayed at rates from 0.56 to 1.12 kg/ha (FMC Corporation, 2001). The presence of at least one boll weevil in a pheromone trap at the perimeter of a field triggers a spray application over the entire field. However, it is desirable to spray only areas that include VC plants as opposed to the entire field, because boll weevils feed only on the VC plants, and Malathion can be dangerous to human health and may also kill beneficial insects when overapplied. The U.S. Environmental Protection Agency classifies Malathion as a toxicity class III pesticide, which has a maximum allowable residue of 8 parts per million (ppm) in food crops like corn and sorghum (United States Environmental Protection Agency, 2016). Spraying only the areas that contain VC plants could help solve the overapplication problem and reduce chemical application costs. For precise spray applications, VC plants need to be accurately detected and located, but doing so is extremely challenging, as the plants tend to be hidden in fields among crop plants and weeds.

Existing practices for detecting VC plants involve field scouting at the perimeters of cotton fields with other current crops (Yadav et al., 2022). This method allows VC plants in the middle of fields to remain undetected and potentially serve as a habitat for boll weevils. Additionally, manual field scouting is time-consuming and tedious, potentially causing delays in reporting and insecticide application. Pimenta et al. (2016) showed that boll weevils prefer cotton's reproductive growth stage, from the first squaring phase which occurs nearly 40 days after germination. Therefore, it is preferable to detect and locate VC plants during the earlier vegetative growth stage and then either destroy them or spray them with insecticide to prevent infestation by boll weevils. Westbrook et al. (2016) showed that aerial remote sensing can be used to detect and locate clusters of VC plants in corn, grain sorghum (milo) and soybean fields. They used linear spectral unmixing on multispectral images of both conventional and okraleaf cotton. Yadav et al. (2019) showed that spectral and textural image features can be used to detect individual VC plants in a corn field. The use of machine learning models trained with textural features resulted in detection accuracies greater than 70%.

Partel et al. (2019) showed that artificial intelligence (AI) tools like machine vision (MV) algorithms that use convolutional neural network (CNN) architectures can be used for near real-time detection of weeds and precisely spraying desired targets or locations. They used edge computing devices involving NVIDIA Pascal (Santa Clara, CA, USA) architecture-based Graphics Processing Units (GPUs) integrated on (1) an NVIDIA GTX 1070 Ti graphics card and then on (2) an NVIDIA Jetson TX2 development board (Santa Clara, CA, USA). The former effort achieved a detection accuracy of 70% on actual plants and 90% on artificial plants, while the latter effort achieved accuracy of 59% on actual plants and 90% on artificial plants. They designed a smart sprayer system to be attached behind an all-terrain vehicle (ATV). Ferreira et al**.** (2017) showed that CNNs can be used to discriminate broadleaf and grass weeds from soybean plants with an accuracy of nearly 98%. They compared detection results of CNNs with those of classical machine learning (ML) techniques like support vector machine (SVM) and Random Forest classification, and CNNs outperformed the classical ML techniques.

With the advent of advanced edge computing devices, real-time image processing has become a reality and is being used in areas like medicine, space, the environment, and agriculture. Shi et al. (2016) noted that UAS for agricultural remote sensing may be superior to other forms of aerial remote sensing like satellite-based remote sensing because of its ability to generate higher spatial and temporal resolution image data. UAS can thus be used for applications like measurement of plant height (Han et al., 2018), predicting yield (Anderson et al., 2019), herbicide spraying, etc. UAS with edge computing devices like GPUs and MV algorithms can potentially be used for near real-time VC detection and precise spot spray applications. Hence, the first objective of this research was to determine whether a popular CNN-based MV model called YOLOv3 can be used for VC detection in a corn field with RGB (red, green, and blue) aerial images collected by a UAS. The second objective was based on the fact that most CNNs are not robust to changes in the scales of input images (Singh & Davis, 2018). Ammar et al. ( 2019) found that the performance of YOLOv3 degraded as the difference in image resolution between training and test data increased. Therefore, to avoid degradation in YOLOv3 performance due to difference in image resolution between training and test datasets, all our experiments were conducted on similar resolutions between the training and test datasets. The purpose was to determine the maximum image size that can be used for VC detection without compromising the performance of the model. This study assumed that using larger images to detect VC plants would save GPU computation time for real-time applications. Therefore, the second objective involved determining the behavior of YOLOv3 on input images at three different scales: 320 x 320 (S1), 416 x 416 (S2), and 512 x 512 (S3) pixels based on three-performance metrics. Those metrics are average precision (AP), mean average precision (mAP) and F1-score at 95% confidence level (α = 0.05).

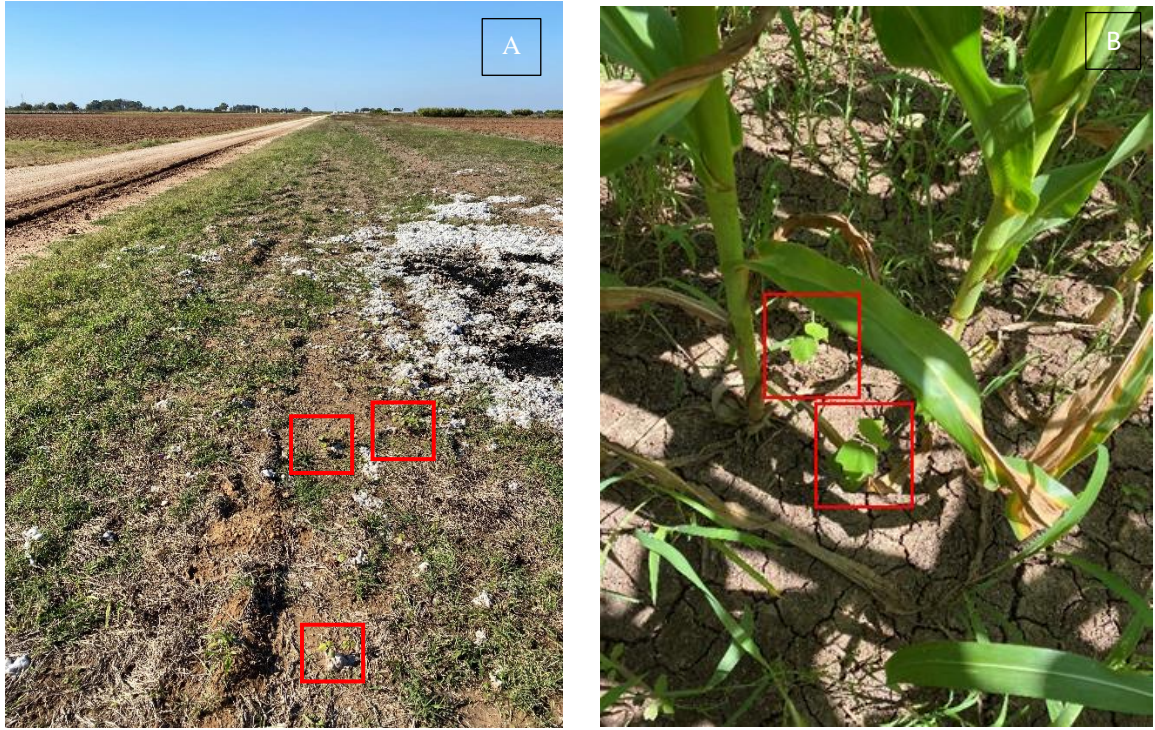

**Fig. 1.** Volunteer cotton (VC) plants seen within red bounding boxes at the perimeter of inter-seasonal cropping field (A). VC plants growing in the middle of inter-seasonal corn field (B).

**2.0 Materials and Methods**

*2.1 YOLOv3 Architecture*

YOLOv3 is state-of-the-art CNN architecture, based on a single neural network (NN) detector, extensively used for real-time object detection in images and videos. The backbone of this CNN architecture is Darknet-53 (Redmon and Farhadi, 2018), which acts as a feature extractor for the model. YOLOv3 is composed of 53 convolution layers in the feature extractor part and another 53 layers in the object detector part, forming a total of 106 layers. Of these, layers 82, 94 and 106 are responsible for detecting objects at three different scales. Large objects are detected by layer 82, medium-size objects by layer 94, and small objects by layer 106. The input is commonly of shape (*n x W x H x c*), where *n* represents number of images, *W* stands for width of images in pixels, *H* is the height of images in pixels, and *c* is the number of channels in the images (in the case of RGB images, red, green, and blue are the three channels). *W* and *H* are in multiples of 32 like 416 x 416, 512 x 512, etc. If an input image is not of this size, the algorithm can resize the image automatically. YOLOv3 has three predefined bounding boxes (also known as anchor boxes) at all the three scales, resulting in a total of nine anchor boxes. In the first step, YOLOv3 divides the input image into grids of 13 x 13, 26 x 26, and 52 x 52 cells. In the second step calculations are performed to determine whether the center point of an object in ground-truth annotated boxes falls in the grid cells. All the grid cells in which the center of the object falls become responsible for predictions of bounding boxes (BB) for detected objects. Then the nine predefined anchor boxes are assigned to these grid cells: i.e., three anchors to each grid cell at all the three scales. Then Darknet-53 is used to extract features of objects from the responsible grid cells. After that, the remaining 53 layers of the detector part are used to predict the BB. Each predicted BB is given by $t_x$, $t_y$, $t_w$ and $t_h$, where $t_x$ and $t_y$ represent central coordinates and $t_w$ and $t_h$ represent the width and height. However, to make sure these values are always positive and between 0 and 1, they are transformed into $b_x$, $b_y$, $b_w$ and $b_h$ with a sigmoid function for $t_x$ and $t_y$ and an exponential function for $t_w$ and $t_h$, as shown in equations (1, 2, 3 and 4).

$$b_x = \sigma(t_x) + c_x \tag{1}$$

$$b_y = \sigma(t_y) + c_y \tag{2}$$

$$b_w = p_w e^{t_w} \tag{3}$$

$$b_h = p_h e^{t_h} \tag{4}$$

Here, $p_w$ and $p_h$ represent the width and height of the predefined anchor boxes assigned to the grid cell, while $c_x$ and $c_y$ represent the top-left corner coordinates of the grid cell. After the BB are predicted, their "objectness" scores are calculated with logistic regression. Values close to 1 represent true positives, while those close to 0 represent false positives. Based on a set threshold value for objectness, most of the predicted BB are filtered out. Then the metric of binary cross entropy is used to determine class prediction for each remaining BB. In the last step, which is the post-processing step for BB, a non-maximum suppression (NMS) algorithm is used to filter out the remaining unwanted BB, leaving only the most appropriate ones. A simplified version of this architecture is shown in Fig. 2 as generated by the visualization software, Netron version 4.8.1 (Kist, 2021).

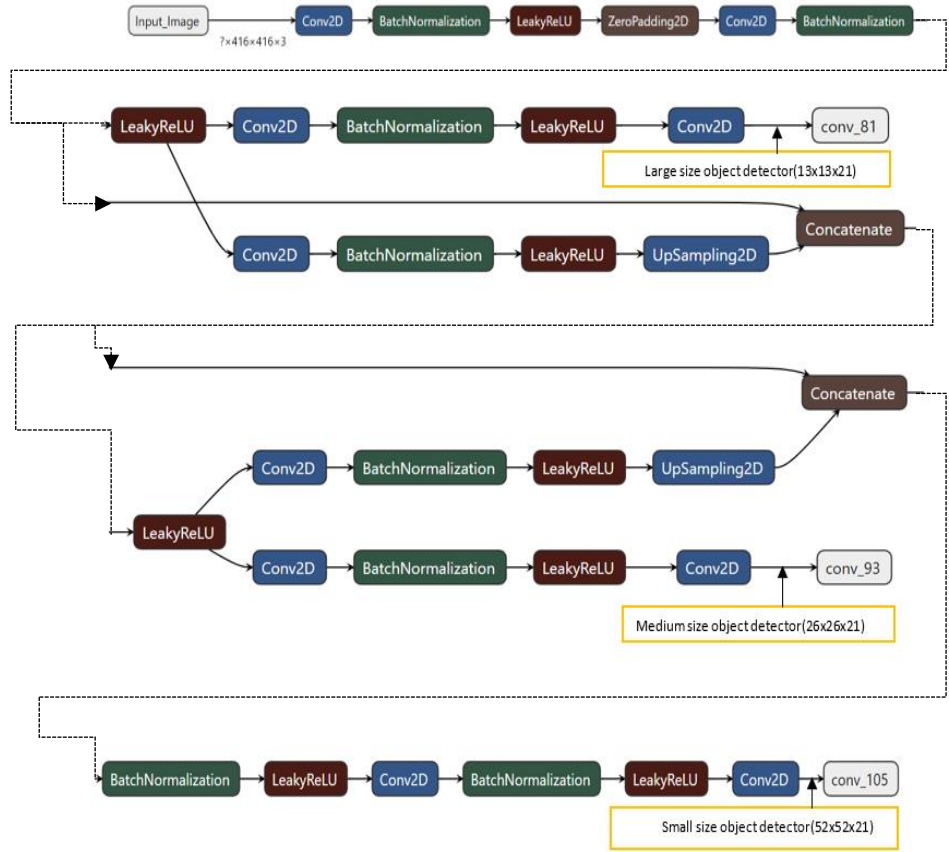

**Fig. 2.** Simplified YOLOv3 architecture visualization generated by Netron visualization software

*2.2 Salient Features of YOLOv3*

As opposed to previous versions, YOLOv3 makes use of three important features: the use of residual block by making multiple skip connections, the use of logistic regression instead of softmax as a classifier, and the use of predefined BB as in Faster Region-based CNN (R-CNN) (Redmon and Farhadi, 2018).

*2.2.1 Residual block*

An improvement that YOLOv3 implements over previous versions is the use of residual blocks by making multiple skip connections between input and output layers. Skipping connections can bypass a block of convolution layers by transferring input features directly from one layer to deeper layers. This helps to solve the problem of vanishing gradient (Veit et al., 2016) by introducing shortcut paths. In CNNs, gradients tend to exponentially decrease during backpropagation. By skipping connections at multiple layers, the residual blocks in YOLOv3 are immune to vanishing gradients because they pass over a shorter route that can allow the gradients to pass from beginning to the end of the network (Veit et al., 2016). A simple four-block residual network is shown in Fig. 3.

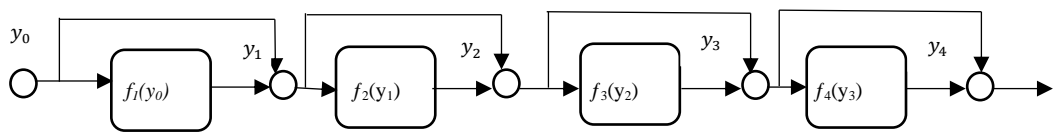

**Fig. 3.** An example of a four-block residual network.

Assuming $y_{i-1}$ as input to an i-th residual block, the output of this block is given as:

$$y_i = f_i(y_{i-1}) + y_{i-1} \tag{5}$$

Here, $f_i(y_{i-1})$ is an output of convolution operations, batch normalization and rectified linear unit (ReLU) activations that can be mathematically written as:

$$f_i(y_{i-1}) \equiv W_i * \sigma(B(W'_i * \sigma(B(y_{i-1})))) \qquad (6)$$

Here, $W_i$ and $W'_i$ are weight matrices, * represents the convolution operation, $B(y_{i-1})$ is the batch normalization operation, where

$$\sigma(B(y_{i-1})) \equiv \max(y_{i-1}, 0) \qquad (7)$$

*2.2.2 Softmax vs. logistic regression*

Softmax is a generalized version of logistic regression that is used as a multi-class classifier (Stanford, 2013) in many CNN architectures including previous versions of YOLO, and it is mathematically written as:

$$s(x_i) = \frac{e^{x_i}}{\sum_{j=1}^{m} e^{x_j}} \qquad (8)$$

As seen in equation (8), softmax transforms all the '*m*' numbers in a feature map into a probability distribution function that ranges from 0 to 1. This classifier works well if every grid cell of an output feature map contains a single class. However, its performance fails when there are multiple classes in a single grid cell of either or all of the three feature maps generated at layers 82, 94 and 106 (Redmon and Farhadi, 2018). This is a common occurrence in many image data sets, thus logistic regression was adopted in YOLOv3 to predict the probability of an object belonging to a particular class making it a multi-label classification as opposed to multi-class classification.

*2.3 Performance Metrics of YOLOv3*

In this study, we used average precision (AP), mean average precision (mAP) and F1-score as the three-performance metrics to evaluate the trained YOLOv3 model. The value used for Intersection over Union (IoU) threshold as explained by Zhou et al. (2019) was 0.30 (30%). Hence, AP is called $AP_{30}$ and mAP is called $mAP_{30}$ in this paper. The general formulae to calculate these metrics are as follows:

$$P = \frac{TP}{TP+FP} \tag{9}$$

$$R = \frac{TP}{TP+FN} \tag{10}$$

$$F1 = \frac{2PR}{P+R} \tag{11}$$

$$AP = \sum_{k=0}^{k=n'-1} [R(k)-R(k+1)]*P(k) \tag{12}$$

$$mAP = \frac{1}{n'}\sum_{k=1}^{k=n'} AP_k \tag{13}$$

Here, *P* represents precision, *R* represents recall, *TP* represents true-positive, *FP* represents false positive, *FN* represents false negative, $AP_k = AP$ of class *k, n'* = number of classes (n'=2 in our case) for mAP and number of thresholds for AP (*n'=1* in our case). This representation means that *R(n') = 0* and *P(n') = 1* for AP, which can also be interpreted as the area enclosed by the precision-recall curve plot and the co-ordinate axes (Zhang et al., 2020).

*2.4 Experiment Site*

The study was conducted at a corn field (Fig. 4; 97°56'21.538"W, 26°9'49.049"N) near Weslaco, Texas, which lies in Hidalgo County in the LRGV, an active eradication zone of TBWEP (Texas Boll Weevil Eradictaion Foundation Inc., 2020). Soil types in the study field are Harlingen clay and Raymondville clay loam (USDA-Natural Resources Conservation Service, 2020). The corn field was planted with 105 cotton seeds of the Phytogen 350 W3FE (CORTEVA agriscience,

Wilmington, Delaware) variety at randomized locations to mimic the presence of VC plants. Some were planted in line with corn beds while others were planted in the furrow middles.

**Fig. 4.** Study area at a corn field near Weslaco, TX

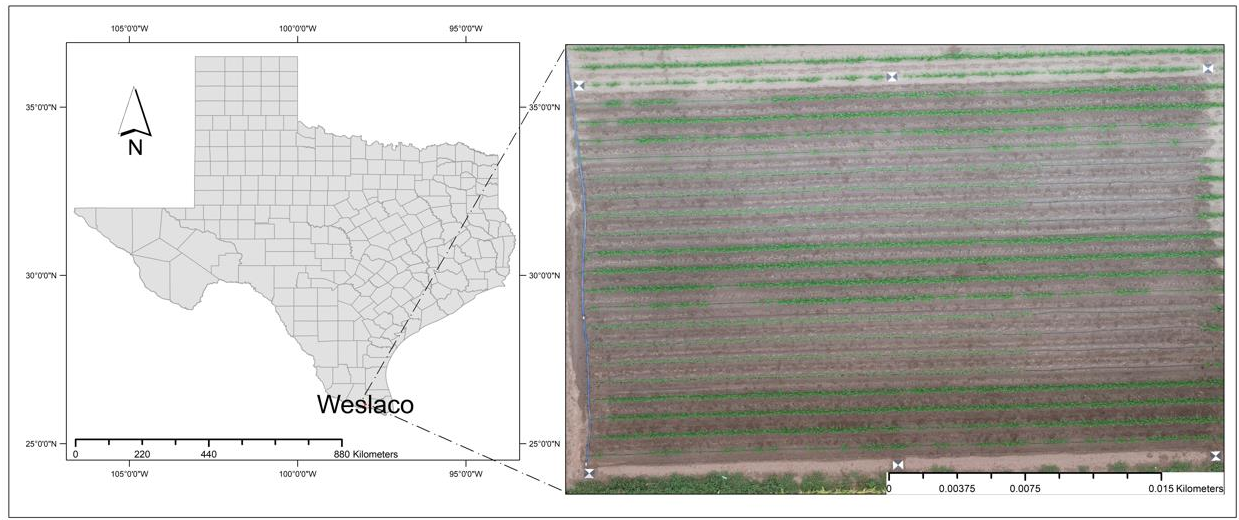

*2.5 Image Data Acquisition*

Fig. 4 shows an orthomosiac image of the test area generated with Pix4DMapper 4.3.33 software (Pix4D S.A., Switzerland). A three band (RGB: red, green, and blue) FC6310 camera (Shenzhen DJI Sciences and Technologies Ltd., Shenzhen, Guangdong, China) integrated on DJI Phantom 4 Pro quadcopter (Shenzhen DJI Sciences and Technologies Ltd., Shenzhen, Guangdong, China) was used to collect aerial images of the test field from 18.3 m (60 ft) above ground level (AGL) altitude. The camera has a resolution of 5472 x 3648 pixels. The resulting images had a ground sampling distance (GSD) of 0.5 cm/pixel (0.20 in./pixel). Data were collected on April 7, 2020, between 10 a.m. and 2 p.m. Central Standard Time (CST). Images were acquired at 80% sidelap and 75% overlap .

*2.6 Data preparation for YOLOV3 model training*

A Python version 3.8.10 (Python Software Foundation, Beaverton, Oregon, USA) script was developed to split images of size 5472 x 3648 pixels into three different sizes 320 x 320, 416 x 416, and 512 x 512 pixels (Yadav, 2021). Five groups of 150 images each were subsequently prepared for each image size. The groups containing images of size 320 x 320 were called S1, 416 x 416 were called S2, and 512 x 512 were called S3. Afterward the LabelImg V-1.8.0 (Tzutalin, 2015) software tool was used to annotate ground-truth bounding boxes for two classes, corn plants and VC plants. Hence, each group (S1, S2 and S3) had 750 images, for a total of 2,250 images. Training and test image datasets were prepared in the ratio of 4:1. This process is represented in the workflow pipeline shown in Fig. 5.

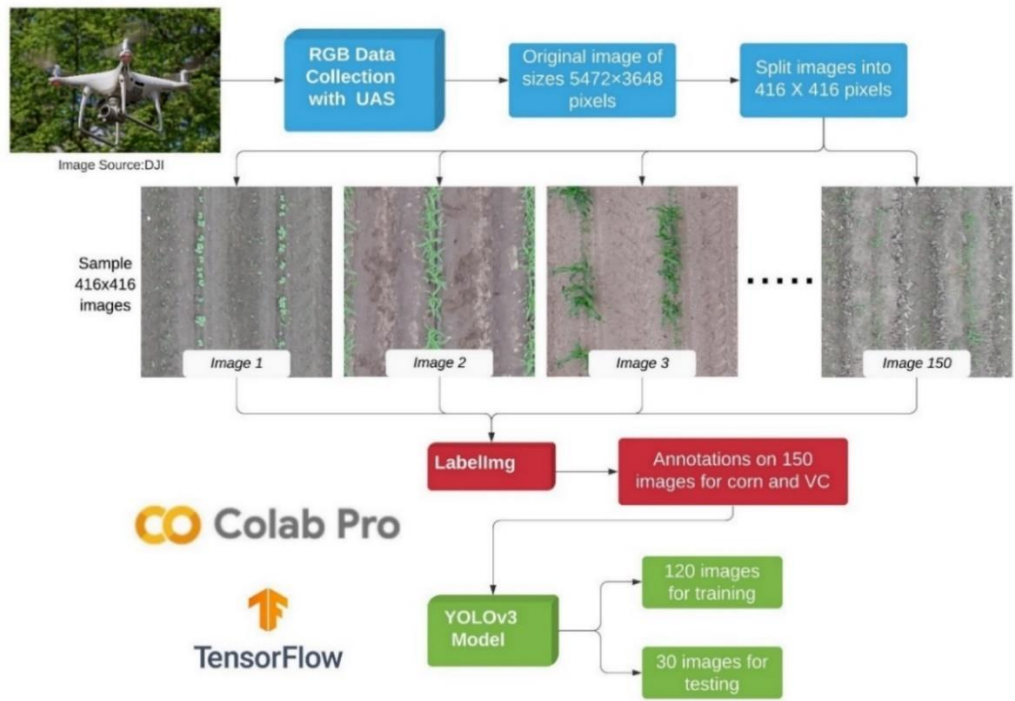

**Fig. 5.** The workflow pipeline showing the processes involved in training and testing YOLOv3 model for volunteer cotton detection in a cornfield.

*2.7 Model Training and Hyperparameters Tuning*

Once the training and test datasets were annotated, the Google Colab Pro (Google LLC., Melno Park, CA) platform was used to train YOLOv3 on an NVIDIA Tesla P100-PCIE GPU (Santa Clara, CA) running Compute Unified Device Architecture (CUDA) version 11.2 and driver version 460.32.03. Code from AlexeyAB's GitHub (GitHub Inc., San Francisco, CA) repository of YOLOv3 was used with some modifications(Alexey et al., 2021). Hyperparameters like learning rate, batch size, train, and test IoU (Intersection over Union) threshold were optimized from different trials and observations based on the convergence of validation loss graphs (Fig. 6).

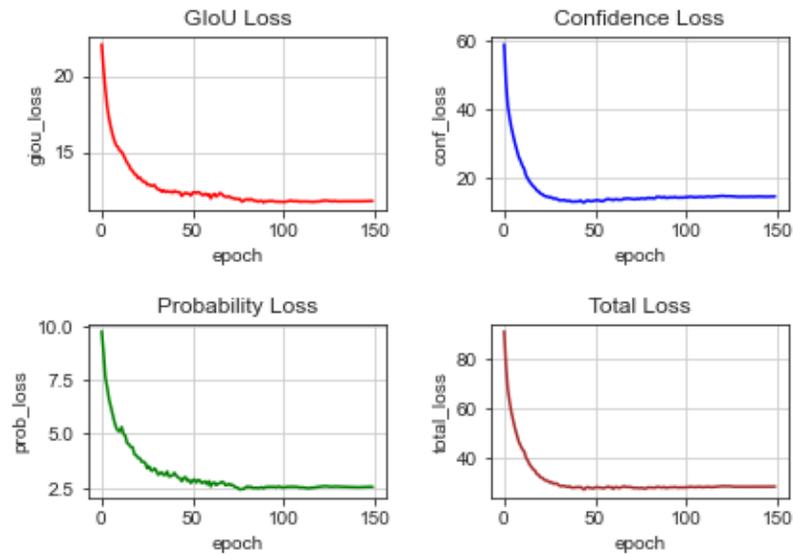

**Fig. 6.** Example of graphs showing different losses during training process.

The YOLOv3 model was trained for a total of 150 iterations (epochs) using 120 images for training and 30 for testing at three different image scales. The graphs in Fig. 6 show losses calculated during the YOLOv3 training process. The bottom right graph, "total loss," shows the sum of the three other losses (confidence loss, GIoU loss and probability loss). Hence, total loss was used to determine the optimal number of training epochs based on its convergence and over-fitting during the training process. Transfer learning approach – i.e., by using the pretrained

ImageNet (Russakovsky et al., 2015) weights as the starting point was used in the training process for faster convergence of the losses and reducing the need for larger training datasets as shown by Jintasuttisak et al.(2022). After multiple trials of training and evaluation, acceptable hyperparameters for this study were found (Table 1).

**Table 1.**

Hyperparameter values that were used in this study.

| Hyperparameters | Values |
| --- | --- |
| Initial learning rate | 0.00001 |
| Final learning rate | 0.0000001 |
| Training batch size | 4 |
| Testing batch size | 2 |
| IOU loss threshold | 0.30 |
| Test score threshold | 0.30 |
| Test IOU threshold | 0.45 |
| Train epoch | 150 |

**Table 2.**

Summary of YOLOv3 model used for the study.

| Summary | Values |
| --- | --- |
| Input Tensor Shape | $S^* \times S^* \times 3$ |
| Total Parameters | 61581727 |
| Total Trainable Parameters | 61529119 |
| Total Non-Trainable Parameters | 52608 |

$S^* \times S^* = 320 \times 320$, $416 \times 416$, $512 \times 512$

Table 2 summarizes the parameters of the entire model and shows the number of trainable and non-trainable parameters that were used in the YOLOv3 architecture.

*2.8 Experiment Design*

A test field was prepared with 20 rows each of approximately 35 m (115 ft) in length and row spacing of approximately 90 cm (3 ft). Twelve rows were planted with sorghum variety F2P134 (Forage Genetics International, Nampa, Idaho) on February 11, 2020, with a seed spacing of 30 cm so that no thinning was required later. Cotton seeds of variety Phytogen 350 W3FE (CORTEVA agriscience, Wilmington, Delaware) were planted on March 9, 2020. The remaining 8 rows were planted with corn of Pioneer variety (CORTEVA agriscience, Wilmington, Delaware) on March 11, 2020, with seed spacing of nearly 30 cm. In the analysis of images, both corn and sorghum were treated as the same class, "corn," as they appeared similar in the early growth phase, and they were treated as a background class wherein the core objective was to detect VC plants. To generate more training data for VC plants, images were collected from an adjacent field where only cotton of the same variety was planted, and the plants were in approximately the same growth phase (between emergence and first square). An image augmentation technique was also used to generate enough training data for both corn and VC plants. With respect to the first objective – identifying VC plants in a corn field – optimal values for hyperparameters were found as shown in Table 1. Then these values were fixed for the second objective – to determine the behavior of YOLOv3 on images at three different scales. YOLOv3 can dynamically adjust input image size by applying down-sampling ratio of 32 thereby making it suitable for detecting objects in images of different sizes. However, in general, the detection accuracy on smaller images is lower compared to that on larger images as seen in a study done by Zheng et al. (Zheng et al., 2020). In their study, they trained the YOLOv3 model with images at three different scales (320 x 320, 384 x 384, and 416 x 416 pixels) to make it more robust. However, in our case, we studied the model's robustness at the three scales S1, S2 and S3 by designing a means and fixed effect experiments in which the three experimental units were $mAP_{30}$, $AP_{30}$ and F1-max. Five observations in each experimental unit were made. S1, S2 and S3 were the three treatments groups. The means model was designed to test whether means of each experimental unit under all the three treatment groups were significantly different. The fixed effect model was designed to test whether any of the treatment groups had a significant effect on the response variable, i.e., the mean of the experimental units. Assuming $Y_{ij}$ as the *j-th* observation from treatment group *i* where *i = S1, S2* and *S3* and *j= 1,2,3,4,5*, the means, and fixed effect models can be expressed in respective order as:

$$Y_{ij}=\mu_i+\varepsilon_{ij} \tag{14}$$

$$Y_{ij}=\mu_.+\alpha_i+\varepsilon_{ij} \tag{15}$$

Here, $\mu_i$ and $\mu_.$ represent means of treatment group *i* and average of $\mu_i$, respectively. $\varepsilon_{ij}$ is the error, or the residual term, associated with both the models. The residual is an independent random variable assumed to have zero mean, constant variance, and normal distribution. $\alpha_i$ is called the *i-th* treatment effect. The null and research or alternative hypotheses for both the models can be expressed as:

$$H_{0M} : \mu_{S1} = \mu_{S2}= \mu_{S3} \tag{16}$$

$$H_{1M} : \mu_{S1} \neq \mu_{S2} \text{ OR } \mu_{S2} \neq \mu_{S3} \text{ OR } \mu_{S1} \neq \mu_{S3} \tag{17}$$

$$H_{0E}: \alpha_{S1} = \alpha_{S2} = \alpha_{S3} \tag{18}$$

$$H_{1E} : \alpha_{S1} \neq \alpha_{S2} \text{ OR } \alpha_{S2} \neq \alpha_{S3} \text{ OR } \alpha_{S1} \neq \alpha_{S3} \tag{19}$$

Where $H_{0M}$ and $H_{1M}$ represent null and alternative hypotheses for the means model, while $H_{0E}$ and $H_{1E}$ represent null and alternative hypotheses for the fixed effect model. To test these models, parametric one-way analysis of variance (ANOVA) was conducted for the normally distributed data with JMP Pro version 15.2.0 software (SAS Institute, North Carolina, U.S.A.), and non-parametric Kruskal-Wallis ANOVA (Wilcoxon Test) was conducted for the data that did not meet the normality assumption because one-way ANOVA may yield inaccurate estimate of $p$-value when data are not normally distributed (Hecke, 2012). The Shapiro-Wilk test (Shapiro and Wilk, 1965) was used to verify the normality assumption at 95% confidence level ($\alpha = 0.05$) using Python's *SciPy Stats* module (The SciPy Community, 2021). Assuming $y_1, y_2, …, y_n$ as the $n$ samples that have normal distribution, the test statistic $W$ for Shapiro-Wilk test is given as:

$$W = \frac{\left(\sum_{i=1}^{n} a_i\ y_i\right)^2}{\sum_{i=1}^{n} (y_i - m_1)^2} \tag{20}$$

Where $m_1$ is the mean of $n$ ordered samples $y_i$ and is given as:

$$m_1 = \frac{1}{n} \sum_{i=1}^{n} y_{(i)} \tag{21}$$

$a_i$ are the normalized coefficients as explained by Shapiro and Wilk (Shapiro and Wilk, 1965).

Based on the outcome of this test, the parametric ANOVA or the non-parametric Wilcoxon Test was performed. After that post-hoc pairwise comparison was done with Tukey's Honestly Significant Difference (HSD) test for the parametric data and Dunn's Test for the non-parametric distribution (Dunn, 1961; Tukey, 1949). The post-hoc analysis was done to determine which treatment group means differed significantly for each experimental unit so that the trained YOLOv3 model would not be applied for input images belonging to the corresponding treatment scale.

## 3.0 Results

*3.1 Overall performance of YOLOv3 for VC detection*

In all the five experimental observations at each of the three scales (i.e., 15 observations in total), the model's total loss was found to converge within the 150 training iterations as seen in Fig. 6. In some cases, it was found to converge around the $50^{th}$ iteration (Fig. 6). Fig. 7 shows histograms along with density plots representing the distributions of $mAP_{30}$, $AP_{30}$ and F1-max for the detected VC plants. To test the likelihood of normal distribution of the observed datasets for further parametric and non-parametric tests, Shapiro-Wilk test was done from which it was found that F1-max was the only normally distributed (parametric) performance metric, while the other two metrics were non-parametric. Fig. 8 shows bar graphs of the performance metrics $mAP_{30}$ and $AP_{30}$ for both corn and VC along with the bar graphs for F1-max and frames per second (FPS). The mean for $mAP_{30}$ was found to be 80.38 with a standard deviation of 9.12 (Fig. 9). The distribution of data can also be seen from box plots in Fig. 9. These results show that the model was trained adequately to detect both the VC and corn classes with an overall accuracy of more than 80%. The mean of $AP_{30}$ for VC was found to be 81.2 with a standard deviation of 12.2 indicating that the model was trained adequately to detect VC plants with a class-specific accuracy of more than 81% in a corn field at all the three treatment levels-S1,S2 and S3. The average F1-max value was found to be approximately 78.5% with standard deviation of 9.35 showing the weighted average detection accuracy of more than 78% at all the three treatment levels-S1,S2 and S3.

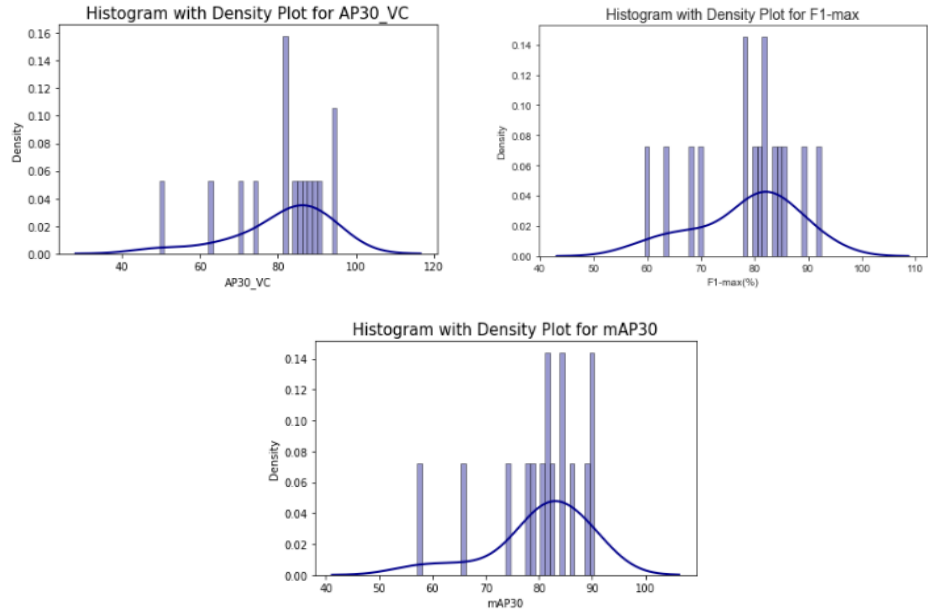

**Fig. 7.** Histograms with density plots showing distribution of mAP$_{30}$, AP$_{30}$ and F1-max for VC plants.

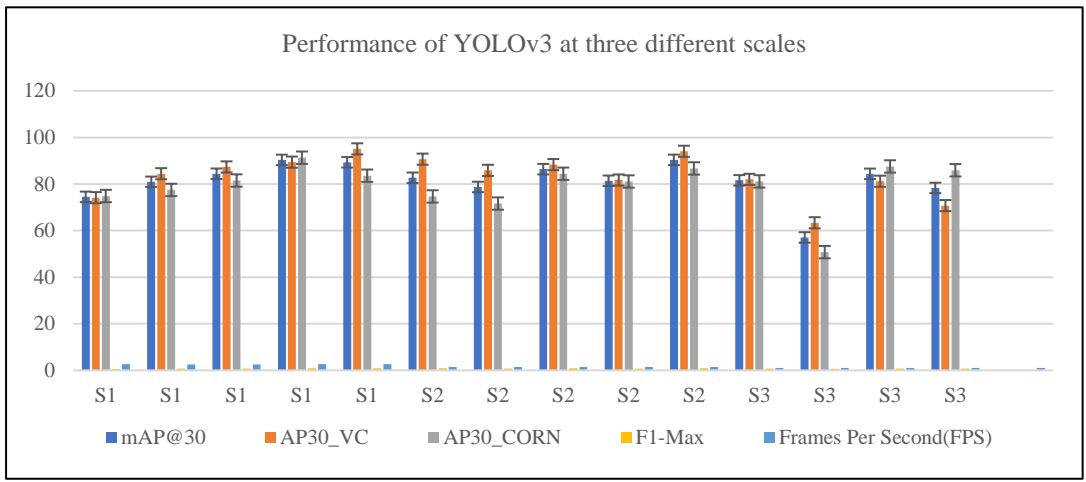

**Fig. 8.** Bar graphs showing the performance of YOLOv3 based on mAP$_{30}$, AP$_{30}$ and F1-max for VC plants along with AP$_{30}$ for corn and detection frames per second (FPS).

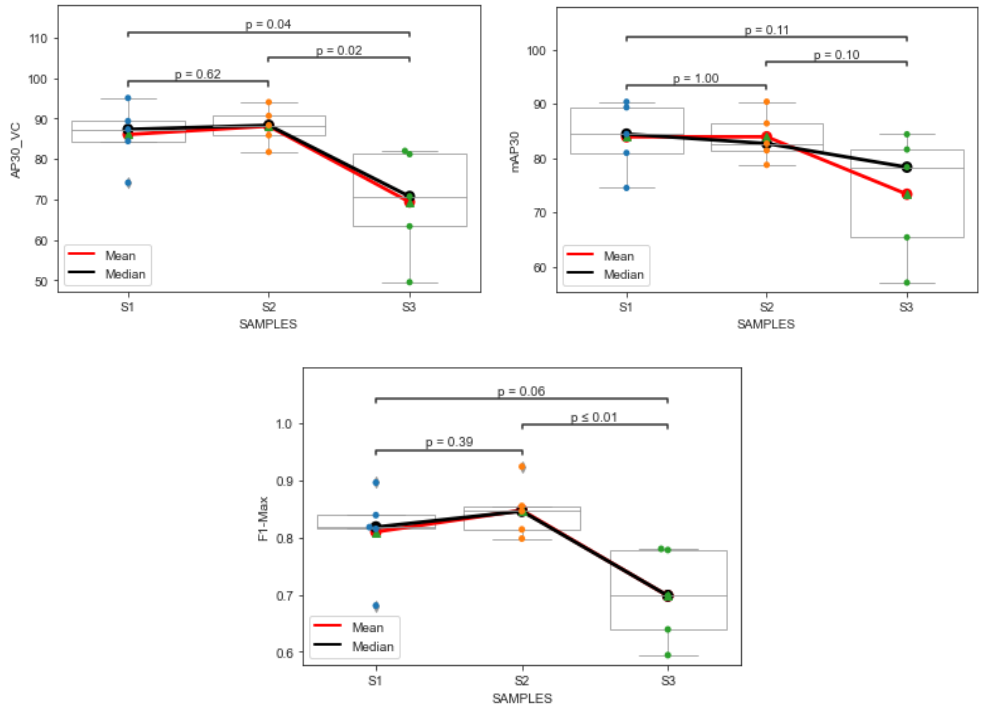

**Fig. 9**. Box plots showing mean and median lines connected for each treatment group. The *p-values* are for significance levels at *α = 0.05*.

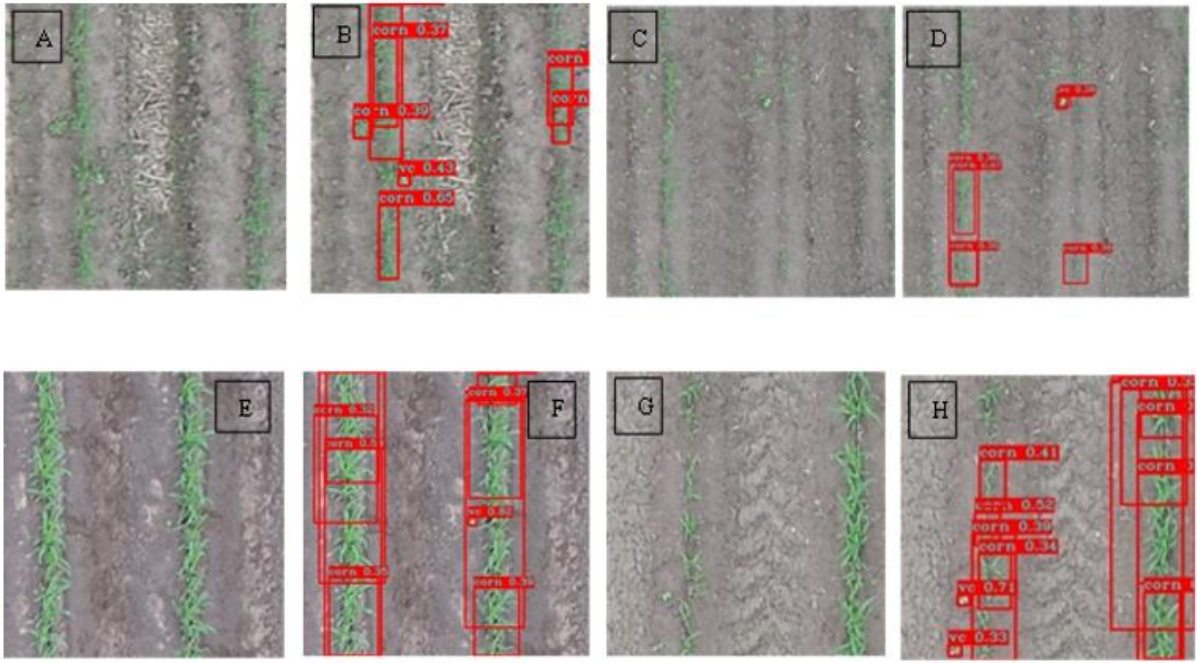
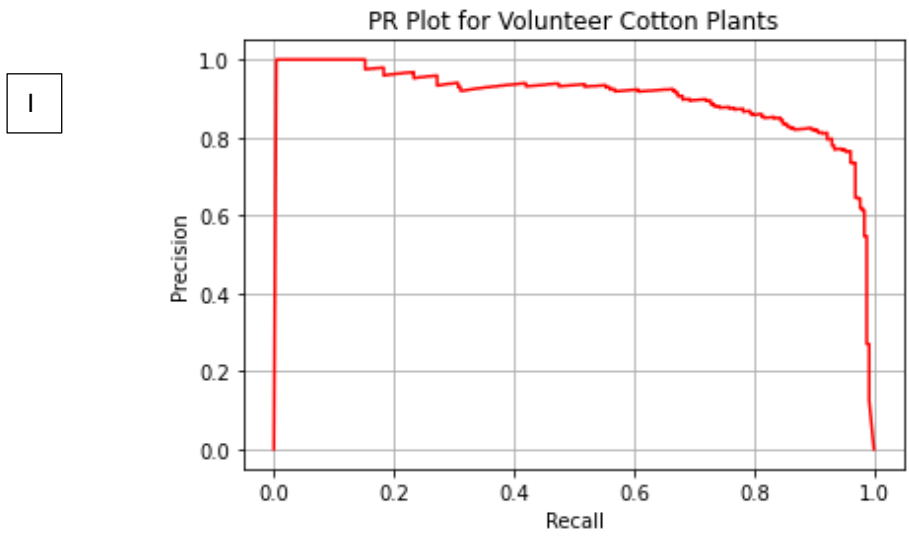

**Fig. 10**. VC plants detection at different regions of a corn field (B, D, F and H). Precision-Recall curve (PRC) obtained as a result during one of the testing instances (I).

Figs. 10-B, D, F and H show detected objects (VC and corn) within red bounding boxes along with their confidence scores. Fig. 10-I shows an example of precision-recall curve (PRC) showing the area under the PRC representing the AP for VC plants.

*3.2 Test of robustness at three treatment levels*

To test if the three group mean values at the three scales (S1, S2 and S3) for F1-max were statistically different i.e., if the three levels of scales (S1, S2 and S3) had any effect on the F1-max values, one-way ANOVA was conducted and the results can be seen in Table 3. The result is significant at 95% confidence level, implying that the three levels of scales had significant effect on the measured F1-max values and $AP_{30}$.

**Table 3.**

Parametric and Non-parametric tests for different performance metrics.

| Metrics | Parametric/Non-Parametric Test | p-value |
|---|---|---|
| F1-Max | One-way ANOVA | 0.0167 |
| AP30 | Kruskal-Wallis (Wilcoxon) | 0.0255 |
| mAP30 | Kruskal-Wallis(Wilcoxon) | 0.1845 |

The Tukey's HSD test showed (Table 4) significant evidence that the means of F1-max were different between the treatment levels S2 and S3 .

**Table 4.**

Tukey's HSD pairwise comparison for F1-max.

| Level | - Level | Difference | Std Err Diff | Lower CL | Upper CL | p-Value |
|---|---|---|---|---|---|---|
| S2 | S3 | 0.1496 | 0.045442 | 0.028373 | 0.270827 | 0.016411 |
| S1 | S3 | 0.11194 | 0.045442 | -0.00929 | 0.233167 | 0.071397 |
| S2 | S1 | 0.03766 | 0.045442 | -0.08357 | 0.158887 | 0.693058 |

These results mean the weighted average of the model's detection accuracy was not consistent between the two scales S2 and S3 but was consistent between the scales S1 and S3 as well as S1 and S2. The non-parametric one-way ANOVA equivalent Kruskal-Wallis (Wilcoxon)

Test for $AP_{30}$ of VC (Table 3) resulted in significant evidence to reject the null hypothesis of equations 16 and 18 implying that group means at the three levels of scales were not equal and therefore the three levels of scales had significant effect on the measured $AP_{30}$ values. The results of Dunn's Test for all pairs with joint ranks to pairwise compare the group means of each treatment level are shown in Table 5.

**Table 5.**

Dunn's test for pairwise comparison of $AP_{30}$ with Bonferroni adjusted p-value.

| Level | - Level | Score Mean Difference | Std Err Diff | Z | p-Value |
|---|---|---|---|---|---|
| S2 | S1 | 0.60000 | 2.828427 | 0.21213 | 1.0000 |
| S3 | S1 | -6.00000 | 2.828427 | -2.12132 | 0.1017 |
| S3 | S2 | -6.80000 | 2.828427 | -2.40416 | 0.0486 |

VC detection accuracy was found to be inconsistent between scales S2 and S3 but was consistent between the other scales. The non-parametric Wilcoxon Test for $mAP_{30}$ did not provide (Table 3) enough evidence to reject the null hypothesis of equation 16. Hence, all the group means of $mAP_{30}$ were similar or not significantly different implying that the three levels of scales didn't have any significant effect on the measured $mAP_{30}$ values.

**4.0 Discussion**

*4.1 VC detection with YOLOv3*

In our previous approach we were able to detect VC plants in a cornfield using classical ML methods and at the side of roads using simple linear iterative clustering (SLIC) Superpixel segmentation method (Wang et al., 2022; Yadav et al., 2019). However, in this paper, we show the application of YOLOv3 deep learning model to detect VC plants (before they reach the pin-head square phase) in an early growth cornfield on aerially collected RGB images at 0.5 cm/pixel GSD. The average detection accuracy (mAP) was higher than 80%, which is over 14% higher than previous results (Yadav et al., 2019). This method of using deep learning as opposed to the traditional machine learning method – i.e., detection once the model was fully trained has the potential to be used for near real-time detection of VC plants in early growth cornfields. By lowering the IoU threshold value to 30%, detection accuracies were found to be higher than at a 50% threshold value. The PRC (Fig. 10-I) shows the relationship between positive predicted values and sensitivity, i.e., fraction of true positives. The focus was to correctly detect maximum number of VC plants, i.e., have the maximum number of true positives (high sensitivity). In this way, the maximum possible number of regions where the boll weevils can survive can be detected and then treated to minimize infestation. Maximizing the specificity (true negatives) can minimize the amount of treatment required, but this was a secondary priority because, in any case, precise treatment based on detection will minimize the amount of chemicals used compared to the broadcast spray of current practice. From previous studies, it has been found that precision as an estimator can be biased towards the classes with larger number of instances i.e. under class imbalance (Zhang et al., 2020). Therefore, in this study, we made sure to have a minimal class imbalance between the VC and corn class instances during the annotation process. The metrics AP and mAP address the issue of precision being biased due to imbalance of class instances (Davis and Goadrich, 2006) and were thus chosen along with the F1-score as performance metrics for the trained YOLOv3 model in our study. Based on this study's findings, the model's performance in detecting the VC plants in a corn field are reliable.

*4.2 Behavior of YOLOv3 on images at three different scales*

The class accuracy ($AP_{30}$) for VC plants shows a slight but non-significant increasing trend from scale S1 to S2 but a significantly decreasing trend from S2 to S3 and S1 to S3. However, the overall accuracy ($mAP_{30}$) does not show a significant change between scales. This difference could possibly mean that presence of corn classes and hence its higher detection accuracy resulted in overall increment of the $mAP_{30}$ values as it is taken into factor for its calculation as shown in equations 18 and 19. The aim of this study was to distinguish between VC and corn, so this result should not be of concern if the sensitivity values for VC are higher. The results show that if the input images belong to the size S1, S2 or S3, the YOLOv3 model will be able to detect VC plants without significantly compromising on performance. In other words, YOLOv3 is robust enough to detect VC plants in a corn field with similar accuracies when the input images belong to any of the three scales.

*4.3 Limitations and recommendations*

In this study, images of the same size as those used for training were tested, limiting the findings to applications only when a model trained with same image size is available. Furthermore, the findings are limited to early growth VC plants, and the results may not hold true for different growth phases of corn and VC plants. To make this study more robust, a similar study could be conducted by training the YOLOv3 model with images of a fixed size and then testing it on images with different sizes. Testing such a trained model on images collected from different AGLs would also be useful.

## 5.0 Conclusions

This studied showed that the deep learning based YOLOv3 model can be used to detect VC plants in an early growth cornfield at a detection accuracy of over 80%, higher than previous texture based classification approaches (Yadav et al., 2019). The study also showed how YOLOv3's performance metrics behave when trained and tested with images of different sizes. Its overall performance remains similar and hence can be used for the three-image sized considered in this study without compromising overall accuracy. This model in general can be used for detecting VC plants in a cornfield at more than 14% higher overall detection accuracy and almost 29% higher detection accuracy at scale S1 than the classical image processing techniques used in previous approach.

## 6.0 Future Direction

In the future it will be helpful to quantify the uncertainty associated with predicted BBs by applying Gaussian modeling to the existing YOLOv3 algorithm so as to improve the localization accuracy, as seen in Choi et al. (2019) and Kraus and Dietmayer (2019). It will also be useful to interpret the trained model's decisions based on VisualBackProp (Bojarski et al., 2016) as well as gradient based saliency map (Simonyan et al., 2014) to visualize what features in the images contributed the most in the model's decision making process. Furthermore, training and testing the model with VC plants after the pin-head square phase will ensure that only boll weevil hostable plants can be detected, and possibly with better accuracy.

**Acknowledgements**

We would like to extend our sincere thanks to all the reviewers, farm manager (Stephen P. Labar) and student assistants (Roy Graves, Madison Hodges, Sam Pyka, Reese Rusk, Raul Sebastian, JT Womack, John Marshall, Katelyn Meszaros, Lane Fisher, and Reagan Smith) involved during field work. This material was made possible, in part, by Cooperative Agreement AP20PPQS&T00C046 from the United States Department of Agriculture's Animal and Plant Health Inspection Service (APHIS). It may not necessarily express APHIS' views.

*CRediT authorship contribution statement*

**Pappu Kumar Yadav**: Conceptualization, Data curation, Formal analysis, Investigation, Methodology, Validation, Visualization, writing – original draft, Writing - review & editing. **J. Alex Thomasson:** Conceptualization, Data curation, Investigation, Methodology, Formal analysis, Project administration, Resources, Supervision, Writing - review & editing. **Robert Hardin:** Project administration, Formal analysis, Resources, Supervision, Writing - review & editing. **Stephen W. Searcy:** Formal analysis ,Writing - review & editing. **Ulisses Braga-Neto:** Investigation, Methodology, Formal analysis ,Writing - review & editing. **Sorin Popescu:** Formal analysis ,Writing - review & editing. **Roberto Rodriguez:** Data curation ,Formal analysis ,Writing - review & editing. **Daniel E Martin:** Writing - review & editing. **Juan Enciso:** Resources, Writing - review & editing. **Karem Meza:** Resources, Writing - review & editing. **Jorge Solorzano Diaz:** Resources, Writing - review & editing. **Tianyi Wang:** Writing - review & editing.

**Declaration of Competing Interest**

The authors declare that they have no known competing financial interests or personal relationships that could have appeared to influence the work reported in this paper.

35
Redmon, J., Farhadi, A., 2018. YOLOv3:An Incremental Improvement, in: Computer Vision and Pattern Recognition. arXiv preprint arXiv:1804.02767, Berlin/Heidelberg, Germany.

Robinson, J.R.C.; Vergara, O., 1999. Structural changes to consider in the valuation of boll weevil eradication programs, in: Proceedings of the Beltwide Cotton Conferences. National Cotton Council. pp. 321–324.

Russakovsky, O., Deng, J., Su, H., Krause, J., Satheesh, S., Ma, S., Huang, Z., Karpathy, A., Khosla, A., Bernstein, M., Berg, A.C., Fei-Fei, L., 2015. ImageNet Large Scale Visual Recognition Challenge. Int. J. Comput. Vis. 115, 211–252. https://doi.org/10.1007/s11263-015-0816-y

Shapiro, A.S.S., Wilk, M.B., 1965. An Analysis of Variance Test for Normality. Biometrika 52, 591–611.

Shi, Y., Alex Thomasson, J., Murray, S.C., Ace Pugh, N., Rooney, W.L., Shafian, S., Rajan, N., Rouze, G., Morgan, C.L.S., Neely, H.L., Rana, A., Bagavathiannan, M. V., Henrickson, J., Bowden, E., Valasek, J., Olsenholler, J., Bishop, M.P., Sheridan, R., Putman, E.B., Popescu, S., Burks, T., Cope, D., Ibrahim, A., McCutchen, B.F., Baltensperger, D.D., Avant, R. V., Vidrine, M., Yang, C., 2016. Unmanned Aerial Vehicles for High-Throughput Phenotyping and Agronomic Research. PLoS One 11, 1–26. https://doi.org/10.1371/journal.pone.0159781

Simonyan, K., Vedaldi, A., Zisserman, A., 2014. Deep inside convolutional networks: Visualising image classification models and saliency maps, in: 2nd International Conference on Learning Representations, ICLR 2014 - Workshop Track Proceedings. pp. 1–8.